\documentclass{article} 
\usepackage{iclr2025_conference,times}

\usepackage{amsmath,amsfonts,bm}

\def\eqref#1{equation~\ref{#1}}

\def\1{\bm{1}}

\DeclareMathAlphabet{\mathsfit}{\encodingdefault}{\sfdefault}{m}{sl}
\SetMathAlphabet{\mathsfit}{bold}{\encodingdefault}{\sfdefault}{bx}{n}

\usepackage{hyperref}
\usepackage{url}

\usepackage{times}
\usepackage{color,xcolor}
\usepackage{epsfig}
\usepackage{graphicx}

\usepackage{adjustbox}
\usepackage{array}
\usepackage{booktabs}
\usepackage{colortbl}
\usepackage{float,wrapfig}
\usepackage{hhline}
\usepackage{multirow}
\usepackage{caption}
\usepackage{amsmath,amsfonts,amsthm,amssymb}
\usepackage{bm}
\usepackage{nicefrac}
\usepackage{microtype}

\usepackage{changepage}
\usepackage{extramarks}
\usepackage{fancyhdr}
\usepackage{lastpage}
\usepackage{setspace}
\usepackage{soul}
\usepackage{xspace}
\usepackage{dsfont}

\usepackage{algorithm}
\usepackage{algpseudocode}
\usepackage{enumerate}
\usepackage{comment}

\usepackage{url}

\definecolor{cvprblue}{rgb}{0.21,0.49,0.74}

\definecolor{nvgreen}{RGB}{118, 185, 0}

\usepackage{pifont}
\usepackage{nccmath}

\usepackage{soul}
\sethlcolor{red}

\usepackage{paralist}

\newcommand{\ignorethis}[1]{}

\makeatletter
\DeclareRobustCommand\onedot{\futurelet\@let@token\@onedot}
\def\@onedot{\ifx\@let@token.\else.\null\fi\xspace}

\def\eg{\emph{e.g}\onedot} 
\makeatother

\makeatletter
\newcommand\footnoteref[1]{\protected@xdef\@thefnmark{\ref{#1}}\@footnotemark}
\makeatother

\newcommand{\vila}[1]{$\text{VILA}_{#1}$}
\newcommand{\mytt}[1]{\texttt{\footnotesize#1}}

\definecolor{nvidiagreen}{rgb}{118,185,0}
\definecolor{mydarkblue}{rgb}{0,0.08,1}
\definecolor{mydarkred}{rgb}{0.8,0.02,0.02}
\definecolor{mydarkorange}{rgb}{0.40,0.2,0.02}
\definecolor{mypurple}{RGB}{111,0,255}
\definecolor{myred}{rgb}{1.0,0.0,0.0}
\definecolor{mygold}{rgb}{0.75,0.6,0.12}
\definecolor{mydarkgray}{rgb}{0.66, 0.66, 0.66}
\definecolor{mydarkgreen}{rgb}{0.02,0.6,0.02}
\definecolor{mygray}{gray}{0.9}
\definecolor{keynotegreen}{rgb}{0.04,0.52,0}
\definecolor{keynoteyellow}{rgb}{1,0.68,0}
\definecolor{mypink}{rgb}{1.0,0.0,0.0}
\definecolor{mylightpurple}{RGB}{243,225,255}
\definecolor{myopen}{RGB}{239	246	237} 
\definecolor{myprop}{RGB}{235	243	250}
\definecolor{mymodel}{RGB}{166,184,150}
\definecolor{visdarkred}{RGB}{209,109,106}
\definecolor{visdarkgreen}{RGB}{157,195,132}
\definecolor{visdarkblue}{RGB}{120,157,229}

\definecolor{visred}{RGB}{251,92,137}
\definecolor{visgreen}{RGB}{124,200,104}
\definecolor{visblue}{RGB}{105,176,241}

\def\methodshort{VILA$^2$\xspace}

\usepackage{wrapfig}
\usepackage{graphicx}
\usepackage[utf8]{inputenc} 
\usepackage[T1]{fontenc}    
\usepackage{hyperref}       
\usepackage{url}            
\usepackage{booktabs}       
\usepackage{amsfonts}       
\usepackage{nicefrac}       
\usepackage{microtype}      
\usepackage{xcolor}         
\usepackage{tcolorbox}
\usepackage{multicol}
\usepackage{tabularx}
\usepackage{makecell}
\usepackage{fixltx2e}
\usepackage{calc}

\title{VILA$^2$: VLM Augmented VLM with Self-Improvement}

\author{Yunhao Fang$^{1}$\thanks{Equal contribution. 
} \quad Ligeng Zhu$^{1}$\footnotemark[1] \quad Yao Lu$^{1}$ \quad Yan Wang$^{1}$ \quad Pavlo Molchanov$^{1}$\\\textbf{Jan Kautz}$^{1}$ \quad \textbf{Jang Hyun Cho}$^{2}$ \quad \textbf{Marco Pavone}$^{1}$ \quad \textbf{Song Han}$^{1,3}$ \quad \textbf{Hongxu Yin}$^{1}$ \\ NVIDIA$^{1}$ \quad UT Austin$^{2}$ \quad MIT$^{3}$
}

\iclrfinalcopy 
\begin{document}

\maketitle

\begin{abstract}

While visual language model (VLM) architectures and training infrastructures advance rapidly, data curation remains under-explored where quantity and quality become a bottleneck. 
Existing work either crawls extra Internet data 
with a loose guarantee of quality or distills from black-box proprietary models (\eg, GPT-4V / Gemini) that are API frequency and performance bounded.
This work enables a VLM to improve itself via data enhancement, exploiting its generative nature. We introduce a simple yet effective VLM augmentation scheme that includes a self-augment step and a specialist-augment step to iteratively improve data quality and hence, model performance. In the self-augment step, the instruction-finetuned VLM recaptions its pretraining caption datasets and then retrains from scratch leveraging refined data. 
Without any expensive human-in-the-loop annotation, we observe improvements in data quality and downstream accuracy boosts with \textit{three} self-augmentation rounds -- a viable \textit{free lunch} to the current VLM training recipe. 
When self-augmentation saturates, we augment the caption diversity by leveraging specialty skills picked up from instruction finetuning. We finetune VLM specialists from the self-augmented VLM with domain-specific experts, including spatial, grounding, and OCR, to fuse task-aware synthetic data into the pretraining stage. Data quality improvements and hallucination reductions are cross-checked by VLM (GPT-4V, Gemini) and human judges. Combining self-augmentation and specialist-augmented training, 
\methodshort consistently improves the accuracy on a wide range of benchmarks over the prior art, producing a reusable pretraining dataset that is 300x more cost-efficient than human labeling.
%


\end{abstract}

\vspace{-1em}
\section{Introduction}

The success of large language models (LLMs)~\citep{raffel2020exploring,dai2019transformer, gpt3, OpenAI_ChatGPT,touvron2023llama, touvron2023llama2,alpaca,vicuna2023,karamcheti2021mistral,chowdhery2022palm,yi,qwen} has offered the cornerstone for cross-modality tasks. Through the alignment of visual encoders to LLMs, visual language models have enabled myriad appealing capabilities to visual tasks, such as instruction following, zero-shot generalization, few-shot in-context learning (ICL), and enhanced world knowledge~\citep{liu2023llava,alayrac2022flamingo,driess2023palm,chen2023pali,li2023blip,fuyu,bai2023qwen,GPT4,zhu2023minigpt,lin2024vila}. The field has progressed rapidly in the past two years, yielding effective alignment training recipes~\citep{driess2023palm,GPT4,lin2024vila} and model architectures~\citep{liu2023llava,alayrac2022flamingo,driess2023palm,chen2023pali,li2023blip}. 

Contrary to the fast-evolving training enhancement, the underlying human-generated datasets and tasks remain simple~\citep{zhu2023multimodal, schuhmann2022laion, kakaobrain2022coyo-700m, sharma2018cc3m}. 
Given the costly nature of VLM  training, most methods are confined with \textit{coarse-quality} \textit{large-scale} captioning image-text pairs (pretraining), followed by \textit{fine-grained} \textit{small-scale} supervised finetuning (SFT). Enhancement of image-text pairs with millions and billions of instances can inevitably impose a huge amount of human effort, and thus not realistic. Recent methods have observed rewarding distillation possibilities from proprietary commercial models like GPT-4V~\citep{openai2023gpt4v} and Gemini~\citep{gemini_2023}. However, the performance is upper bounded by these models. 
In the meantime, studies remain very sparse on how to better utilize VLMs to correct human error and remedy dataset task simplicity for enhanced training.

\begin{figure}[t]
    \centering
    \vspace{-2em}
    \includegraphics[width=1.0\linewidth]{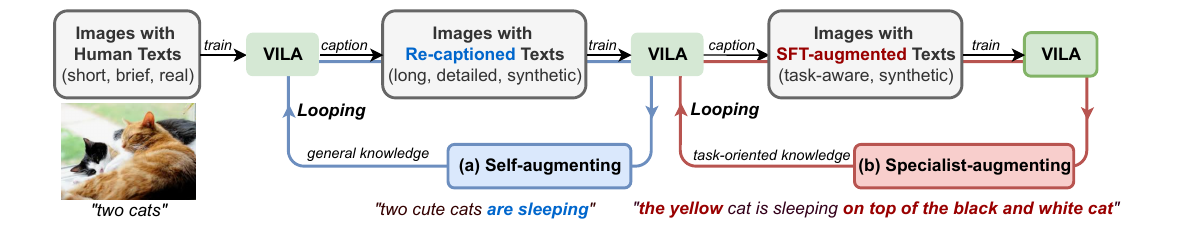}
    \caption{The schematic diagram to train \methodshort, short for VILA-augmented VILA. We re-formulate visual language model (VLM) training with \textit{``model in the loop''} to remedy training data defficacy. We start with validating design options in constructing a self-augmenting loop (Section.~\ref{sec:step1}) to improve on caption quality of the default training task. After the saturation of this process, we challenge the VLM to generate data conforming to extra SFT-enabled tasks to further VLM learning (Section.~\ref{sec:step2}). Our new design insights yield off-the-shelf performance boosts to VLMs (Section.~\ref{sec:exp}). 
    }
    \label{fig:overview}
\end{figure}

In this work, we aim to answer ``\textit{whether it is possible that the VLM itself can remedy dataset deficiency and enhance its training.}'' We delve deep into the potential of using VLM itself to refine and augment pretraining data and performance iteratively. Our new training regime, summarized in Figure~\ref{fig:overview}, consists of two main steps: a \textbf{self-augment step} and a \textbf{specialist-augment step}. 
We start with the self-augment loop \big(Figure~\ref{fig:overview} (a)\big) that leverages VLMs to enhance the quality of pretraining data. We demonstrate that synthetic data, combined with the original data, can collaboratively generate stronger models in a bootstrapped loop manner. Intuitively and as we observed, the loops offer performance boosts \textit{for free}, but suffers diminishing returns after \textit{\textbf{3}} rounds. To facilitate further learning, we reformulate a more challenging task-specific loop \big(Figure~\ref{fig:overview} (b)\big). In this loop, specialists with a focus on new knowledge or tasks, such as a spatial-aware expert, OCR expert or grounding expert, are finetuned from the self-augmented VLM using a limited amount of additional SFT data. The specialists can then recaption a massive amount of pretraining data. Finally, the self-augmented VLM can be retrained on the specialist-recaptioned pretraining data to further boost the performance. 

The insights yield a novel VLM augmentation training regime progressively improves data quality, by transferring knowledge from the \textbf{higher-quality but small-scale SFT} stage back to the \textbf{larger-scale but coarse pretraining} phase. This improvements cover enhanced visual semantics (Figure~\ref{fig:overview}) and reducing hallucinations (Table~\ref{tab:vsr_results}-~\ref{tab:human_eval}). We also observed consistent agreements on data quality improvements when cross checking the data by VLM models (GPT-4V, Gemini) and humans (Ph.D. students).
This offers a direct performance boost to VLMs. We introduce a new family of \methodshort models, as in VLM-augmented-VLM. \methodshort outperforms state-of-the-art methods across main benchmarks, all enhanced without bells and whistles via self-bootstrapped training. 
We hope that the insights and release of \methodshort's recipe, data, and code can assist with our community for better understanding and usage of synthetic data to train stronger VLMs. 

\section{Methodology}

In this paper, we focus on auto-regressive VLMs where image tokens are projected into the textual space and concatenated with text tokens, in line with~\citep{liu2023llava,gemini_2023,lin2024vila}. This approach is chosen because of its flexibility when handling multi-modal inputs. We follow the widely adapted three-stage training paradigm, \textit{i.e.}, align-pretrain-SFT, to ablate our studies. We start to self-augment VLM training by constructing a bootstrapped loop leveraging VLM's general captioning capability. After the bootstrapping saturates, we then introduce specialist augmenting exploiting VLM's skills picked up during SFT across additional visual tasks as specialist feedback to its pretraining stage. We next elaborate on these steps in detail.

\subsection{Self-augmenting via General Knowledge Enhancement}
\label{sec:step1}

Existing VLM training largely relies on data from the Internet, where the texts are usually brief and short, see Table~\ref{tab:recaption_stat} where the average number of words is less than 20 for MMC4~\citep{zhu2023multimodal} and COYO~\citep{kakaobrain2022coyo-700m}. In addition to brevity, human annotations can also fall short in explaining to LLMs the versatile semantics an image presents. As another example, Figure.~\ref{fig:boostraping_viz} indicates that an original COYO caption that only describes a person riding on the street, omitting details about clothing and surroundings. Previous studies have either assigned humans to write dense captions or by using commercial propriety APIs for detailed descriptions. The first option can be labor-intensive and costly, while the second risks model biases, limiting model performance, and potentially raising copyright concerns. 

\begin{figure}[t]
    \centering
    \includegraphics[width=0.95\linewidth]{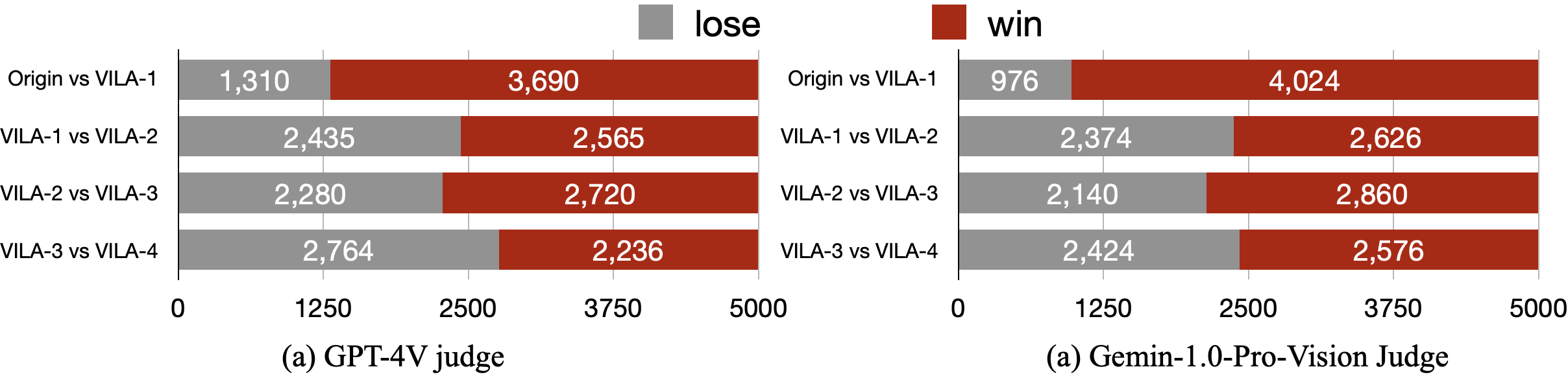}
    \caption{LLM judgement for captions from \vila{i}, $i$ indicating the self-augmenting round. Evaluations are based on 5,000 sampled data from Coyo~\cite{kakaobrain2022coyo-700m}. Both GPT-4V~\cite{openai2023gpt4v} and Gemini-1.0-pro~\cite{gemini_2023} prefer for \methodshort augmented texts and captions from later rounds got higher scores. 
    }
    \label{fig:winrate}
\end{figure}
\begin{table}[t]
    \centering
    \scalebox{0.80}{
    \begin{tabular}{l|ccccccc}\toprule
    &MMC4 &COYO &COYO-\vila{1} &COYO-\vila{2} &COYO-\vila{3} &COYO-\vila{4} \\ \midrule
    Avg \#Words &17.1 $\pm$ 25.0 &11.9 $\pm$ 9.0 &101.2 $\pm$ 43.0 &117.1 $\pm$ 49.1 &126.77 $\pm$ 50.10 &125.9 $\pm$ 51.2 \\ 
    $\text{VQA}^{\text{v2}}$ &  N.A. & 61.6 & 62.5 & 63.5 & \textbf{63.7} & \ul{63.6} \\ 
    \bottomrule
    \end{tabular}
    }
    \caption{The average number of words and $\text{VQA}^{\text{v2}}$ evaluation comparison between the original dataset and the re-captioned dataset. Best performance is bolded and second best is underlined. During self-augmentation, the caption lengths increases significantly, thus offering more details.
    }
    \label{tab:recaption_stat}
\end{table}

Rather than distilling proprietary models or relying on manual laboring, we aim to use \textit{VILA to generate better captions for VILA's pretraining}. This approach exploits the power of the already-intelligent VILA within intermediate training stages to conduct laborious relabelling efforts.  
We begin with the original dataset to train the initial version of VILA, referred to as \vila{0} in subsequent experiments. Next, we use \vila{0} to re-caption VILA's pretraining datasets. With appropriate prompt choice and conversation template, ~\vila{0} is able to generate \textit{long} and \textit{detailed} captions. Then the augmented datasets, consisting of real images from the internet and synthetic texts from ~\vila{0}, are used to train the next round of VILA, named \vila{1}. This self-augmenting process can be repeated several rounds before convergence, leading to detailed descriptions (higher $\text{VQA}^{\text{v2}}$ score in Table~\ref{tab:recaption_stat}) and improved text quality (LLM Judge in Figure.~\ref{fig:winrate}).

\begin{table*}[t]
\setlength{\tabcolsep}{3pt}
\centering
\scalebox{0.8}{
\begin{tabular}{l|c|ccccccccc}\toprule
 & Avg \#words &VQA$^\text{v2}$&GQA &SQA$^I$ &VQA$^\text{T}$ & POPE &LLaVA$^\text{W}$& MM-Vet&MMMU  \\ \midrule
Baseline & 17.1 & 79.6 &62.4&68.4&61.6&84.2&68.4&34.5&33.8 \\ \midrule
\rowcolor{black!10} \multicolumn{10}{l}{\textit{Prompt Ablation for Self-Augmenting}}\\
Self-augment Iter1 - Simple & 90.4 & 79.4 &\ul{63.0} &68.7 &\ul{62.4} &\textbf{87.0} &68.3 &34.5 &33.1 \\
Self-augment Iter1 - Long v1 & 94.8 &80.0 &62.7 &\textbf{71.1} &62.2 &84.0 &71.7 &\ul{34.5} &{34.4} \\ 
Self-augment Iter1 - Long v2 & 105.4 & \textbf{80.1} & 63.2 &70.7 &62.7 &84.6 &71.7 &34.9 & \ul{34.7} \\
Self-augment Iter1 - Long v3 & 102.4 & \textbf{80.1} & 63.4 &71.0 &62.9 &85.0 &71.4 &34.4 &34.7 \\
\midrule
\rowcolor{black!10} \multicolumn{10}{l}{\textit{Conversation Template Ablation for Self-Augmenting}}\\
Mixed - re-caption text only & 101.2 & 79.6 &62.5 &\textbf{71.1} &62.3 &81.0 & \ul{71.8} &34.2 &34.1 \\
\rowcolor{myopen}  Mixed - concatenated & 127.3 & 80.0&\textbf{63.2} &71.0 & \textbf{62.5} & \ul{85.0} &\textbf{72.2} & \textbf{34.8} &\textbf{35.8} \\
\bottomrule
\end{tabular}
}
\caption{
    Comparison with different prompts and training templates when self-augmenting for one round. The best and second-best results are highlighted with \textbf{bold} and \ul{underline} respectively. The results show that prompts design are critical for self-augmenting. Re-captioning the dataset with naive prompt "\textit{Describe the image briefly}" does not improve while designed prompt can significantly boost the mode performance. 
}
\label{tab:prompt_exp_comparison}
\end{table*}
\subsubsection{Prompts and Template Design}


The choice of prompt is particularly important for immediate performance improvements. To validate prompt design choices, we conducted an in-depth study on prompt choices as follows and discuss our findings, where \mytt{<img>} indicates the location where image features will be inserted,
{\footnotesize
\begin{itemize}
    \item {Prompt Simple}: \mytt{<img> Describe the image briefly.}
    \item {Prompt Long-v1}:\mytt{<img> Describe the image in details.}
    \item {Prompt Long-v2}:\mytt{<img> Elaborate on the visual and narrative
elements of the image in detail.}
    \item {Prompt Long-v3}: \mytt{<img> Instead of describing the imaginary content, only describing the content one can determine confidently from the image. Do not describe the contents by itemizing them in list form. Minimize aesthetic descriptions as much as possible.}
\end{itemize}
}

\textbf{Brief and Short Re-captioning is NOT Helpful.}
We begin with a straightforward prompt asking VLMs to \textit{briefly} describe the image. Despite these brief recaptions being significantly longer than the original texts (90 vs. 17 words), there is no notable improvement in VLM benchmarks, as shown in Table~\ref{tab:prompt_exp_comparison}. In fact, metrics even deteriorate in benchmarks such as Science-Image and MMMU-Test. This decline may stem from a lack of details during recaption.

Next, we redesign the prompt to encourage VLMs to provide a more detailed description of visual narrative elements in images. We also referenced the prompt template from ShareGPT-4V~\citep{chen2023sharegpt4v} to ensure the descriptions are accurate and precise. Our experiments demonstrate that using three different long prompts improves the quality of recaptioning and boosts performance in benchmarks, detailed in Table~\ref{tab:recaption_stat}. Therefore, we leverage a mixture of these prompts by randomly selecting from versions v1 to v3.

\textbf{Keeping Original Human Text is Important.}
We compare different conversation templates in Table~\ref{tab:recaption_stat}. The first template uses only real human data, while the ``concatenated" approach adapts both human and synthetic descriptions. Our experiments reveal that using self-augmented data improves performance on major benchmarks like LLaVA-Bench, Science-Image, TextVQA. However, we noticed a decline in several metrics. This prompted us to concatenate both the original and re-captioned texts to best preserve information, which consistently improves all VLM metrics.
(Table~\ref{tab:recaption_stat}).


\subsection{Surpassing the Limit with Specialist VLM Augmentation}
\label{sec:step2}

While self-augmentation provides a simple yet effective way to boost VLM's performance, we notice that the improvement starts to saturate with all free lunches having been squeezed (Table~\ref{tab:iteration_rounds}). We hypothesize that this shortcoming stems from the monotonic task of \textit{general} descriptive captioning, which is also heavily influenced by language modeling priors.

To advance the boundary of self-augmentation, we propose the integration of extra \textit{task-specific} knowledge into a generalist VLM to create several specialist VLMs. Each specialist model is finetuned with data that demands a deeper understanding of image components and semantics, \eg, spatial relations, localization, and OCR. A bootstrapped loop can then transfer such specialist knowledge from small-scale SFT data onto a large number of pretraining images.

\subsubsection{Acquiring Specialized Knowledge}
\begin{figure}[h]
    \centering
    \includegraphics[width=0.95\linewidth]{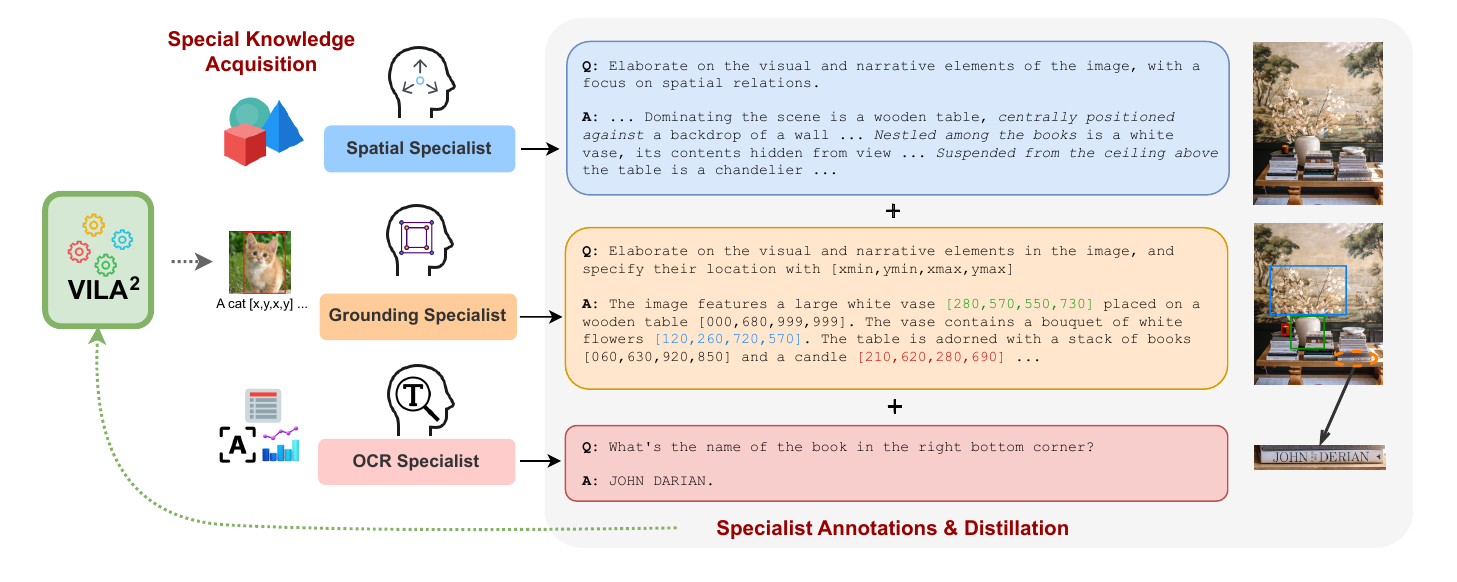}
    \caption{\methodshort specialist VLM augmentation overview. 
    We gather task-specific knowledge to create task-specialist VLMs. These specialist VLMs annotate images with task-oriented prompts and generate question-answering pairs to re-train the next iteration of \methodshort.}
    \label{fig:specialist_schematic}
\end{figure}

We focus on three challenging tasks: \textit{spatial relations understanding}, \textit{grounded narration}, and \textit{OCR} (Figure~\ref{fig:specialist_schematic}), and then elaborate the specialist construction as follows: 

\textbf{Spatial Specialist.} To explicitly challenge the model to acquire additional spatial awareness, we curated \textit{SpatialRelationQA}, a dataset containing 1 million conversations about spatial relations within scenes. Our dataset is built on LV3D, a comprehensive collection of both indoor and outdoor 3D datasets from Cube-LLM~\citep{cho2024language} that is designed to enhance the understanding of 3D spatial relations requiring both perceptual and grounding skills.

We formulated a two-step process to create the QA pairs.
\begin{enumerate}
    \item For each cleaned 3D scene, we iterated through all 3D bounding boxes and randomly sampled from  object-object relations (\mytt{closest, in front of, behind, left, right}) and object-camera relations (\mytt{close, far, closest, farthest, left, right});
    \item Next we checked if any remaining bounding boxes matched these sampled relations. For each matched results, we randomly selected question templates to construct the QA pairs, incorporating instances, their projected 2D bounding boxes, and relations.
\end{enumerate}
A sample question can be constructed as: ``\textit{Where is the \textit{chair closest to} the \textit{table [$x_{\text{left}}$,$y_{\text{top}}$,$x_{\text{right}}$,$y_{\text{bottom}}$]} in the image?}'', with answer being the target bounding box. During recaptioning, we guide the model to answer with more spatial information by prompting the specialist VLM with evenly sampled templates during the recaptioning phase, e.g., ``\textit{<image> Can you explain the content of the image and their spatial relations in detail?}''.

\textbf{Grounding Specialist.} To enhance knowledge of grounding awareness, we exploited grounded narration -- a highly visual-centric task that requires VLMs to generate detailed captions to accurately locate major visual elements using 2D bounding boxes, as shown in Figure~\ref{fig:specialist_schematic}. This approach provides dense supervision and allows us to verify if VLMs hallucinate.
To develop the \methodshort grounding specialist, we used image-grounded caption pairs from the 20M GRIT dataset~\citep{peng2023kosmos}. We first filtered out bounding boxes covering more than 70\% of the image area, as many images in GRIT are book or album covers unrelated to the captions. We then removed images containing more than three instances of the same category to reduce complexity and decrease noise in generation orders. This process yielded 4M high-quality instances for grounding specialist training, which we split into two subsets: \textit{Grounding-Short} (3M) and \textit{Grounding-Long} (838K) for two-stage finetuning.

\textbf{OCR Specialist.} For OCR capabilities, we utilize a diverse set of images featuring textual content, such as tables, charts, and documents, to develop an OCR specialist. Each image is annotated with QA pairs that focus on text recognition (\eg, \textit{Q: What is the title of the book?}), comprehension (\eg, \textit{Q: Which bar has the largest value?}), and reasoning (\eg, \textit{Q: What is the main idea of the quote from Albert Camus?}). Dataset details are provided in appendix~\ref{sec:appendix_A_2}.

The three specialist are then applied to the final augmentation stage. We use a new set of task-oriented prompts to activate the specialists' knowledge and improve their instruction-following ability by narrowing focus. Specifically, during the specialist augmentation stage.  we prompt with evenly sampled templates of \mytt{"<image> Elaborate on the visual and narrative elements of the image in detail, with a focus on spatial relations." }and \mytt{"<image> Can you explain the content of the image and their spatial relations in detail?"} 
Similarly, the grounding specialist generates captions with bounding boxes for the major visual focus, while the OCR specialist identifies most textual content in the images. The responses from these different specialist VLMs are appended to the original captions  as QA pairs for the next pretraining iteration of \methodshort. The full details are attached in appendix~\ref{sec:appendix_A_1_1}.

\vspace{-1em}
\section{Experiments}
\label{sec:exp}

\textbf{Model Architecture.} We follow the architecture from VILA~\citep{lin2024vila}, where a multi-modal large model consists of three key components: an \textit{LLM} for auto-regressive generation, a \textit{visual encoder} for extracting visual features, and an image-text \textit{projector} to align the modalities.

We use Llama2-7B~\citep{touvron2023llama2} for exploratory experiments to address the question ``\textit{To what extent can a VLM self-bootstrap?}''. Then we switch to our previous SOTA settings with Llama3-8B-Instruct~\citep{AI@Meta} and Yi-34B~\citep{yi} when compared to other methods. For visual encoders, we use SigLIP~\citep{zhai2023sigmoid} for LLaMA-series models and InternViT-6B~\citep{chen2023internvl} for the Yi-34B model. For projection layers, we follow LLaVA~\citep{liu2023llava, liu2024llavanext} to adapt simple linear layers for bridging image and text modalities. At the same time, we introduce a 4$\times$ downsampling of visual tokens by concatenating $2\times2$ neighboring patches along the channel dimension and using a simple MLP for the downsampling process.

\begin{figure}[t]
    \centering
    \includegraphics[width=0.90\linewidth]{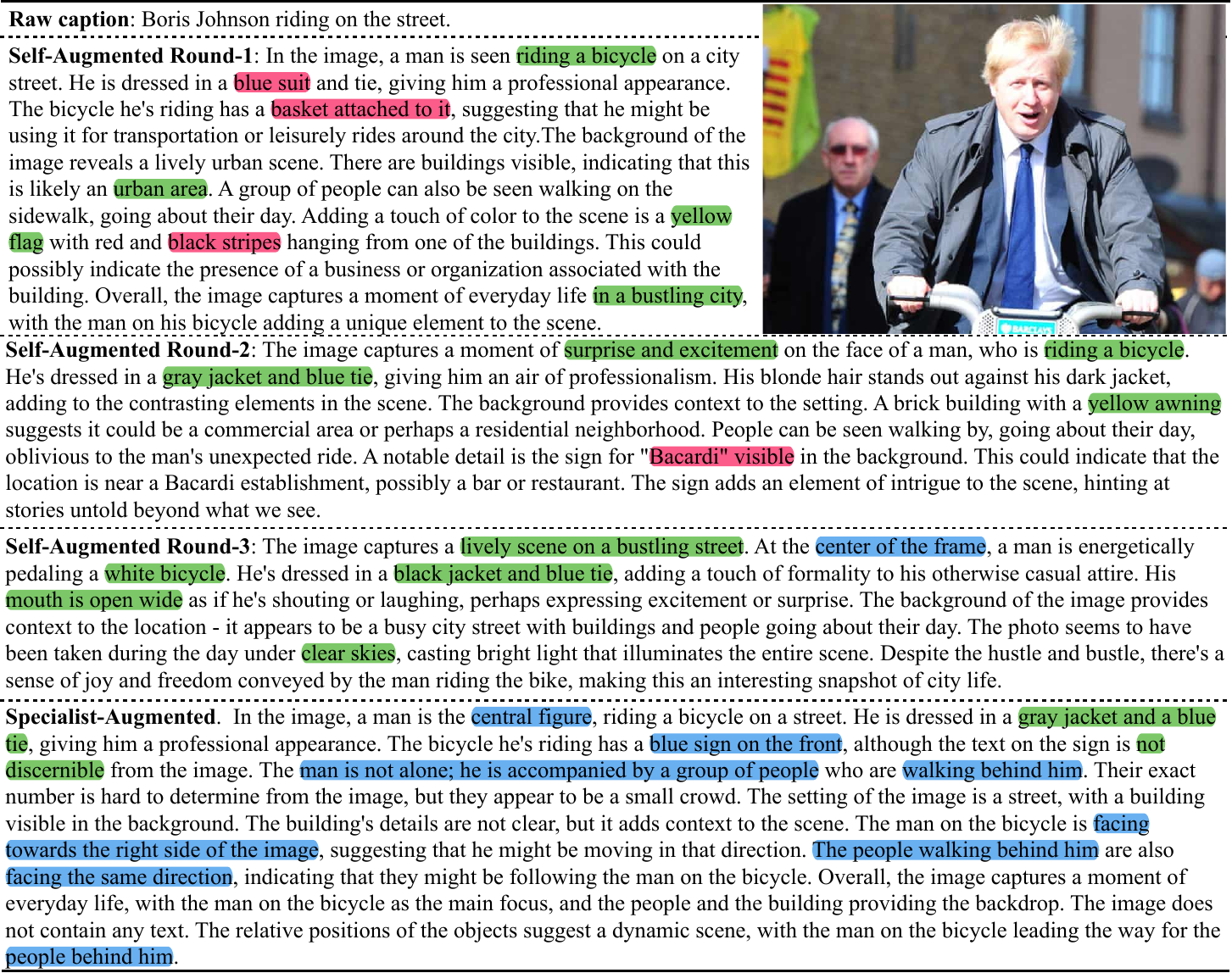}
    \caption{
         \methodshort continuously enhances caption quality over self-augmenting. The sample is from the COYO. We mark facts in \colorbox{visgreen}{green}, hallucinations in \colorbox{visred}{red}, and spatial information in \colorbox{visblue}{blue}. (Please zoom out for the best viewing experience)
    }
    \label{fig:boostraping_viz}
\end{figure}

\textbf{Training Strategies.} We conduct \methodshort training following widely used three-stage settings.

\vspace{-0.5em}
\begin{enumerate}
    \setlength{\itemsep}{0em}
    \item \textit{Projector Initialization}. The language models and ViT are separately pretrained, while the projector is randomly initialized. To initially align the feature space between the visual and text modalities, we utilize the LLaVA align dataset~\citep{liu2023llava}.
    \item \textit{Vision-Language Pretraining}. We then pretrain the model (LLM and the projector) on the visual language corpus. 
    We consider interleaved image text corpus (\eg, MMC4~\citep{zhu2023multimodal}) and image-text pairs (\eg, COYO~\citep{kakaobrain2022coyo-700m}). \textbf{We apply our \methodshort for the pretraining data} and the augmented data will be applied in this stage to replace original COYO captions.
    \item \textit{Visual Instruction-tuning}. Finally, we perform instruction tuning of the pretrained model on visual language instruction datasets. The blending details is attached in the appendix. 
\end{enumerate}
\vspace{-0.5em}
Without specifically mentioned, our experiments are conducted with 128 GPUs and a global batch size of 1024. We employ AdamW optimizer with learning rate \{$10^{-3}$, $5\times10^{-5}$, $2\times10^{-5}$\} for aforementioned three stages respectively. Each stage is trained with one epoch with a 3\% warmup strategy. No weight decay is applied. In some self-/specialist augmented training, \methodshort may involve extra stage to further improve. Please refer to Section.~\ref{sec:step2} and Appendix~\ref{appendix:training_detail} for more details.

\textbf{Data}.  
Our pretraining stage consists of 6M sampled MMC4~\citep{zhu2023multimodal}, 25M sampled Coyo~\citep{kakaobrain2022coyo-700m}, and the full ShareGPT4V~\citep{chen2023sharegpt4v}. To ensure a fair comparison, we only replace the text captions during our experiments while keeping all image sources the same. We use two SFT data blends for different purposes: a smaller blend of 1.8M samples for exploratory experiments in Table~\ref{tab:iteration_rounds}--Table~\ref{tab:specialist_comparison}; a larger blend of 5.9M samples augmented from VILA's training receipt, for SOTA experiments in Table~\ref{tab:sota_comparison}--Table~\ref{tab:mmmu_comparison}. Detailed SFT recipe and specialist data fulllist can be found in the Appendix.~\ref{appendix:sft_data}-Appendix.~\ref{appendix:specialist_data}.

\textbf{Evaluation}. We ablate our models in the following common VLM benchmarks. Note that some metrics are shortened due to space limits. 
VQA$^\text{v2}$~\citep{goyal2017vqav2}; GQA~\citep{hudson2019gqa}; SQA: ScienceQA~\citep{lu2022learn}; VQA$^\text{T}$: TextVQA~\citep{singh2019textvqa}; POPE~\citep{li2023pope}; MMB: MMBench~\citep{liu2023mmbench}; MMB$^\text{CN}$: MMBench-Chinese~\citep{liu2023mmbench}; SEED: SEED-Bench~\citep{li2023seed}; LLaVA$^\text{W}$: LLaVA-Bench (In-the-Wild)~\citep{liu2023llava}; MM-Vet~\citep{yu2023mmvet}; MMMU~\citep{yue2023mmmu}. 

\subsection{Self-Augmentation Results}
Our goal is to "augment" existing pretraining datasets by rewriting captions with dense and informative texts. Instead of relying on human labor or black-box APIs, we use VILA to generate these captions. Therefore, the enriched caption can help develop better VILAs based on which VILA can also feedback to further enhance the captions for the training dataset.

\textbf{\methodshort Enriches Dataset Text Quality.}
The main VLM's performance boost stems from improved caption quality. As shown in Table~\ref{tab:recaption_stat}, caption length increases rapidly after self-augmentation and plateaus around rounds 3 and 4. This aligns with the trend observed in the benchmark results (round 1: 12 to 101, round 3: 117 to 126). 
Though Caption length does not increase significantly after round-1, we continue to observe consistent improvement on VLM benchmarks. We hypothesize that self-augmentation beyond round-1 primarily enhances caption quality by providing more accurate details and reducing hallucination, as visualized in Figure~\ref{fig:boostraping_viz}. The initial brief caption is brief and short (only describing Boris’s riding action). The later captions evolves to include more details about clothing and surroundings. Although early iterations may contain some hallucinations (such as a non-existent basket and misread "Barcardi" text), subsequent iterations refine the caption to include only visual elements that can be confidently identified (more evidence in Table~\ref{tab:vsr_results}-Table~\ref{tab:human_eval}).

\textbf{\methodshort Bootstraps VLM's Performance.}
We follow the same pretraining + SFT process as VILA~\citep{lin2024vila} and sample 5\% data from the pretraining phase to ablate. The images are from the existing COYO~\citep{kakaobrain2022coyo-700m} and MMC4~\citep{zhu2023multimodal} and in each loop, we use the models trained last round to generate new captions for half of the sampled COYO images. MMC4 is not re-captioned because of its interleaved feature. Other settings are kept the same. We compare \vila{i} from different looping steps on common VLM benchmarks. We notice that self-augmented data help improves the model performance across different iterations: \vila{i+1} is consistently better than \vila{i} and the looping progressively boosts the performance (\vila{1-4} in Table~\ref{tab:iteration_rounds}). 
The self-augmenting technique is consistently useful until three rounds. \vila{4} reaches saturation and no longer bring consistent improvement of \vila{3}. 

\begin{table*}[t]
\setlength{\tabcolsep}{3pt}
\centering
\vspace{-1em}
\scriptsize
\begin{tabular}{l|cccccccc}\toprule
&VQA$^\text{v2}$&GQA &SQA$^\text{I}$ &VQA$^\text{T}$ & POPE &LLaVA$^\text{W}$& MM-Vet&MMMU \\ \midrule
\vila{0} - Baseline & 79.6 & 62.4&68.4&61.6&84.2&68.4&34.5&33.8 \\ \midrule
\vila{1} &80.0&63.2&71.0 &62.5 &84.6 &72.2 &34.8 &\ul{35.8} \\
\vila{2} &\ul{80.8}&\textbf{63.5}& {71.5} & 63.5&{84.7} &71.2& 34.9 &35.2 \\
\rowcolor{myopen}  \vila{3} & 80.7 &\textbf{63.5}&\ul{71.5}& \ul{63.7} & 84.5 & \textbf{72.3} & {35.5} & {35.5} \\
\vila{4} & {80.7} & 63.4&71.2 & {63.6} & \textbf{85.0} &\textbf{72.3} & \ul{35.5} & 35.0 \\ 
\midrule
\rowcolor{myopen}  \vila{3}+Spatial Specialist & \textbf{81.1} & 62.8 & \textbf{72.9} & \textbf{65.0} & \textbf{85.0} & 71.4 & \textbf{37.1} & \textbf{36.8 }\\
\bottomrule
\end{tabular}
\caption{
    Self-augmenting can consistently enhance the performance of model training. For \vila{1-4}, the best and second-best results are highlighted in \textbf{bold} and \ul{underline}, respectively. 
    With each iteration, VLM improves the quality of the pretraining dataset's captions. These improved descriptions lead to progressively better performance when training subsequent VLMs. Although the effects of self-augmentation plateau after three rounds, they can be further improved by our specialist.
}
\label{tab:iteration_rounds}
\end{table*}

\begin{table*}[b]
\setlength{\tabcolsep}{3pt}
\centering
\scriptsize
\begin{tabular}{l|llllllll}\toprule
&VQA$^\text{v2}$&GQA &VQA$^\text{T}$ & POPE &SEED-I& MME &MM-Vet&MMMU \\ \midrule
\rowcolor{black!10} \multicolumn{9}{l}{\textit{Pretrain Data: 10\%\ MMC4-core+10\%\ COYO-25M+ShareGPT4V-Pretrain}}\\
Original Caption& 81.4\phantom{\textsubscript{\textcolor{mydarkgreen}{$\uparrow$0.0}}} & 63.8\phantom{\textsubscript{\textcolor{mydarkgreen}{$\uparrow$0.0}}}& 65.2\phantom{\textsubscript{\textcolor{mydarkgreen}{$\uparrow$0.0}}} & 85.5\phantom{\textsubscript{\textcolor{mydarkgreen}{$\uparrow$0.0}}} & 70.6\phantom{\textsubscript{\textcolor{mydarkgreen}{$\uparrow$0.0}}} & 1472.5\phantom{\textsubscript{\textcolor{mydarkgreen}{$\uparrow$00.0}}} & 34.0\phantom{\textsubscript{\textcolor{mydarkgreen}{$\uparrow$0.0}}}& 31.8\phantom{\textsubscript{\textcolor{mydarkgreen}{$\uparrow$0.0}}} \\
+ Spatial Specialist     & \textbf{81.9}\textsubscript{\textcolor{mydarkgreen}{$\uparrow$0.5}} & \textbf{64.1}\textsubscript{\textcolor{mydarkgreen}{$\uparrow$0.3}} & \textbf{66.0}\textsubscript{\textcolor{mydarkgreen}{$\uparrow$0.8}} & 85.9\textsubscript{\textcolor{mydarkgreen}{$\uparrow$0.4}} & 71.8\textsubscript{\textcolor{mydarkgreen}{$\uparrow$1.2}} & 1476.5\textsubscript{\textcolor{mydarkgreen}{$\uparrow$4.0}}\phantom{\textsubscript{0}} & 36.7\textsubscript{\textcolor{mydarkgreen}{$\uparrow$2.7}} & 32.5\textsubscript{\textcolor{mydarkgreen}{$\uparrow$0.7}} \\
 + OCR Specialist     & 81.8\textsubscript{\textcolor{mydarkgreen}{$\uparrow$0.4}} & 64.0\textsubscript{\textcolor{mydarkgreen}{$\uparrow$0.2}} &  65.3\textsubscript{\textcolor{mydarkgreen}{$\uparrow$0.1}} & 86.4\ \textsubscript{\textcolor{mydarkgreen}{$\uparrow$0.9}} & \textbf{72.1}\textsubscript{\textcolor{mydarkgreen}{$\uparrow$1.5}} & 1500.2\textsubscript{\textcolor{mydarkgreen}{$\uparrow$27.7}} & 34.3\textsubscript{\textcolor{mydarkgreen}{$\uparrow$0.3}} & 32.1\textsubscript{\textcolor{mydarkgreen}{$\uparrow$0.3}} \\
 + Grounding Specialist     & 81.8\textsubscript{\textcolor{mydarkgreen}{$\uparrow$0.4}} & 64.0\textsubscript{\textcolor{mydarkgreen}{$\uparrow$0.2}} &  65.1\textsubscript{\textcolor{mydarkred}{$\downarrow$0.1}} & \textbf{86.7}\textsubscript{\textcolor{mydarkgreen}{$\uparrow$1.2}} & 71.0\textsubscript{\textcolor{mydarkgreen}{$\uparrow$0.4}} & \textbf{1536.4}\textsubscript{\textcolor{mydarkgreen}{$\uparrow$63.9}} & \textbf{37.5}\textsubscript{\textcolor{mydarkgreen}{$\uparrow$3.5}} & \textbf{32.6}\textsubscript{\textcolor{mydarkgreen}{$\uparrow$0.8}} \\ \midrule
\rowcolor{black!10} \multicolumn{9}{l}{\textit{Pretrain Data: MMC4-core+COYO-25M+ShareGPT4V-Pretrain}}\\
Original Caption & 82.2 & 63.9 & 66.7 & \textbf{86.5} & 71.2 & 1518.2 & 42.6 & 33.4 \\ 
+ All 3 Specialists & \textbf{83.0}\textsubscript{\textcolor{mydarkgreen}{+0.8}} & \textbf{64.7}\textsubscript{\textcolor{mydarkgreen}{+0.8}} & \textbf{70.9}\textsubscript{\textcolor{mydarkgreen}{+4.2}} & 86.4\textsubscript{\textcolor{mydarkred}{-0.1}} & \textbf{74.0}\textsubscript{\textcolor{mydarkgreen}{+2.8}} & \textbf{1656.2} \textsubscript{\textcolor{mydarkgreen}{+142} }& \textbf{44.7} \textsubscript{\textcolor{mydarkgreen}{+2.1}} & \textbf{35.8}\textsubscript{\textcolor{mydarkgreen}{+2.4}} \\ 
\bottomrule
\end{tabular}
\caption{
    Effectiveness of the data re-captioned by  specialists:  
    We mark the best performance with \textbf{bold}. The results in the same block are trained with different pretraining data but the same SFT data. 
    Specialists-annotated data consistently improves  on both 10\% and 100\% pretraining setting.
}
\label{tab:specialist_comparison}
\end{table*}

\begin{table*}[t]
\setlength{\tabcolsep}{3pt}
\centering
\scalebox{0.65}{
\begin{tabular}{l ll | lllll | lllll }
\toprule
Method & LLM & Res. & VQA$^\text{v2}$ & GQA & VizWiz &  SQA$^\text{I}$ & VQA$^\text{T}$& MMB & MMB$^\text{CN}$ & SEED & LLaVA$^\text{W}$ & MM-Vet \\
\midrule
BLIP-2~\citep{li2023blip} & Vicuna-13B & 224 & 41.0 & 41 & 19.6 & 61 & 42.5 &  -- & -- & 46.4 & 38.1 & 22.4 \\
InstructBLIP~\citep{Dai2023InstructBLIP} & Vicuna-7B & 224& -- & 49.2 & 34.5 & 60.5 & 50.1  & 36 & 23.7 & 53.4 & 60.9 & 26.2 \\
InstructBLIP~\citep{Dai2023InstructBLIP} & Vicuna-13B & 224 & -- & 49.5 & 33.4 & 63.1 & 50.7 & -- & -- & -- & 58.2 & 25.6 \\
Qwen-VL~\citep{bai2023qwen} & Qwen-7B & 448 & 78.8 & 59.3 & 35.2& 67.1 & 63.8 & 38.2 & 7.4 & 56.3 & -- & -- \\
Qwen-VL-Chat~\citep{bai2023qwen} & Qwen-7B & 448 & 78.2 & 57.5 & 38.9 & 68.2 & 61.5 & 60.6 & 56.7 & 58.2 & -- & -- \\
LLaVA-1.5~\citep{liu2023improved} & Vicuna-1.5-7B & 336 & 78.5 & 62.0 & 50.0 & 66.8 & 58.2 & 64.3 & 58.3 & 58.6 & 63.4 & 30.5 \\ 
LLaVA-1.5~\citep{liu2023improved} & Vicuna-1.5-13B & 336& 80.0 & 63.3 & 53.6 & 71.6 & 61.3 & 67.7 & 63.6 & 61.6 & 70.7 & 35.4 \\


VILA-7B~\citep{lin2024vila} & Llama 2-7B & 336& 79.9 & 62.3 & 57.8 & 68.2 & 64.4& 68.9 & 61.7 & 61.1 & 69.7 & 34.9 \\
VILA-13B~\citep{lin2024vila} & Llama 2-13B & 336& \ul{80.8} & 63.3 & \ul{60.6} & 73.7 & 66.6 & 70.3 & \ul{64.3} & \ul{62.8} & 73.0 & \ul{38.8} \\

LLaVA-NeXT-8B~\citep{liu2024llavanext} & Llama 3-8B & 672 & -- & \textbf{65.2} & -- & 72.8 & 64.6 & 72.1 & -- & -- & \ul{80.1} & -- \\
Cambrian-1-8B~\citep{tong2024cambrian1fullyopenvisioncentric} & Llama 3-8B & 1024 & -- & \ul{64.6} & -- & \ul{80.4} & \ul{71.7} & \ul{75.9} & -- & -- & -- & -- \\
Mini-Gemini-HD-8B~\citep{li2024minigeminiminingpotentialmultimodality} & Llama 3-8B & 1536 & -- & 64.5 & -- & 75.1 & 70.2 & 72.7 & -- & -- & -- & -- \\
\midrule
\rowcolor{myopen}\methodshort-8B (ours) & Llama 3-8B & 384& \textbf{82.9} & 64.1 & \textbf{64.3} & \textbf{87.6} & \textbf{73.4} & \textbf{76.6} & \textbf{71.7} & \textbf{66.1} & \textbf{86.6} & \textbf{50.0} \\

\bottomrule
\end{tabular}
}
\caption{
Comparison with state-of-the-art methods on 10 visual-language benchmarks. Our models consistently improve VILA under a head-to-head comparison, showing the effectiveness of enhanced pretraining data quality. We mark the best performance \textbf{bold} and the second-best \underline{underlined}.
}
\label{tab:sota_comparison}
\end{table*}

\subsection{Specialist Augmentation Results}

The ``self-augmentation then training'' cycle reaches a plateau after three iterations, as illustrated in Table~\ref{tab:iteration_rounds}. However, by incorporating tasks-specific specialists, we can overcome the limit and introduce further performance improvement.

\textbf{Surpassing the Limit with \methodshort Specialist.}
The caption augmented by the specialist (the last example) retains the most visible details and provides more information about spatial relations compared to the "Round-4" caption (Figure~\ref{fig:boostraping_viz}). This additional information includes object-to-object relations, as well as localization and clear pose of the major focus, which are not present in the \textit{SpatialRelationQA} dataset. We hypothesize this improvement might result from combining knowledge in specialist data and VLM's training data. The effectiveness of these enriched captions is demonstrated in Table~\ref{tab:iteration_rounds} (\vila{3} + \textit{spatial specialist}). Following the same SFT stage, we observe notable improvements in 5 out of the 8 benchmarks.

\textbf{The More Specialists, The Better The Performance.}
We next explore the significance of scaling up our \methodshort specialists with more pretraining data using recent state-of-the-art settings. We demonstrate the effectiveness of each specialist using S2~\citep{shi2024needlargervisionmodels} with Llama 3-8B-Instruct~\citep{llama3}.
On a 10\% subset of pretraining data, specialist VLMs show overall improvements across most VQA benchmarks in Table~\ref{tab:specialist_comparison}'s first part. We then combine annotations from all three specialists into multi-round QA pairs for each image and retrain VILA. This synergy among the specialists proves highly effective, with scaling up to the full pretraining set yielding significant improvements. Results on larger models will be discussed in next section.

\subsection{Benchmark Comparison to Prior Work}

We now perform a comprehensive comparison to prior work over 10 major benchmarks and summarize results in Table~\ref{tab:sota_comparison} and Table~\ref{tab:mmmu_comparison}. Note that we used a total of 25 million COYO data sampled from the 700 million with the highest CLIP score. We augment the original short real labels with multi-round QA pairs annotated by three specialists, all from 8B models. For 40B models, we continue to train from the stage 2 checkpoints with a mix of 3.75 M recaptioned COYO and a 200K caption dataset~\citep{chen2024allavaharnessinggpt4vsynthesizeddata}. We observed the improvements in quality is consistent and can scale to 40B VILA checkpoints. The detailed training recipes of our 8B and 40B checkpoints are included in the Appendix and will be released jointly with the code base.

Remarkably we observed the enhanced self-augmentation and specialist augmentation training recipes, backed by enhanced and refined datasets, support \methodshort to further push the performance boundary of VILA~\citep{lin2024vila} by noticeable margins across almost all benchmarks, consistent with the ablated performance boosts we observed in previous analysis of Table~\ref{tab:iteration_rounds}. Moreover, \methodshort now constitutes a SOTA performance on the main MMMU~\citep{yue2023mmmu} test dataset leaderboard across all open-sourced models, without the usage of proprietary datasets and only based on publicly available datasets.
\begin{table*}[ht]
\setlength{\tabcolsep}{3pt}
\centering
\scriptsize
\begin{tabular}{l | c | c c c c c c }
\toprule
Method & \textbf{Overall} (Test/Val) & Art \& Design & Business & Science & Health & Human \& Social & Tech. \& Eng. \\ 
\midrule
\rowcolor{myprop} GPT-4V~\citep{GPT4}      & 56.1 / 56.8  & 65.3 & 64.3 & 48.4 & 63.5 & 76.3 & 41.7 \\
\rowcolor{myprop} SenseChat-V~\citep{sensechatv} & 50.3 / 54.6 & 62.7 & 44.1 & 42.3 & 55.7 & 74.7 & 43.5  \\
\rowcolor{myopen} \textbf{\textcolor{mydarkgreen}{\methodshort-40B (ours)}} & \textbf{\textcolor{mydarkgreen}{47.9 / 53.0}}   & \textbf{\textcolor{mydarkgreen}{62.0}} & \textbf{\textcolor{mydarkgreen}{42.3}} & \textbf{\textcolor{mydarkgreen}{38.5}} & \textbf{\textcolor{mydarkgreen}{51.9}} & \textbf{\textcolor{mydarkgreen}{71.9}} & \textbf{\textcolor{mydarkgreen}{42.3}}  \\
\rowcolor{myprop} Qwen-VL-MAX~\citep{qwen}  & 46.8 / 51.4 & 64.2 & 39.8 & 36.3 & 52.5 & 70.4 & 40.7   \\
\rowcolor{myopen} InternVL-Chat-V1.2~\citep{chen2023internvl} & 46.2 / 51.6& 62.5 & 37.6 & 37.9 & 49.7 & 70.1 & 40.8 \\
\rowcolor{myopen} Cambrian-1-34B~\citep{liu2024llavanext}& \makebox[\widthof{46.8}][r]{-} / 49.7 & - & - & - & - & - & -  \\
\rowcolor{myopen} LLaVA-1.6~\citep{liu2024llavanext}& 44.7 / 48.1 & 58.6 & 39.9 & 36.0 & 51.2 & 70.2 & 36.3 \\
\rowcolor{myopen} Mini-Gemini-HD-34B~\citep{liu2024llavanext}& \makebox[\widthof{46.8}][r]{-} / 48.0 & - & - & - & - & - & -  \\
\rowcolor{myprop} Marco-VL-Plus*          & 44.3 / 46.2 & 57.4 & 34.7 & 38.5 & 48.7 & 72.2 & 36.7  \\
\rowcolor{myopen} Yi-VL~\citep{yi}           & 41.6 / 45.9 & 56.1 & 33.3 & 32.9 & 45.9 & 66.5 & 36.0  \\
\rowcolor{myprop} Qwen-VL-PLUS~\citep{qwen} & 40.8 / 45.2 & 59.9 & 34.5 & 32.8 & 43.7 & 65.5 & 32.9   \\
\rowcolor{myprop} Marco-VL-Plus*           & 40.4 / 41.2 & 56.5 & 31.0 & 31.0 & 46.9 & 66.5 & 33.8  \\
\rowcolor{myprop} Weitu-VL-1.0*            & 38.4 / - & 56.6 & 30.5 & 31.1 & 38.4 & 59.0 & 34.2   \\
\rowcolor{myopen} \textcolor{mydarkgreen}{\textbf{\methodshort-8B (ours)}} & \textbf{\textcolor{mydarkgreen}{38.3 / 40.8}}& \textbf{\textcolor{mydarkgreen}{54.3}} & \textbf{\textcolor{mydarkgreen}{32.0}} & \textbf{\textcolor{mydarkgreen}{29.3}} & \textbf{\textcolor{mydarkgreen}{39.7}} & \textbf{\textcolor{mydarkgreen}{56.8}} & \textbf{\textcolor{mydarkgreen}{34.4}}  \\
\rowcolor{myopen} InfiMM-Zephyr~\citep{InfiMM}& 35.5 / 39.4 & 50.0 & 29.6 & 28.2 & 37.5 & 54.6 & 31.1  \\
\rowcolor{myopen} SVIT~\citep{zhao2023svit}& 34.1 / 38.0 & 48.9 & 28.0 & 26.8 & 35.5 & 50.9 & 28.8\\
\rowcolor{myopen} Emu2-Chat~\citep{Emu2}& 34.1 / 36.3 & 50.6 & 27.7 & 28.0 & 32.4 & 50.3 & 31.3  \\
\rowcolor{myopen} BLIP-2 FLAN-T5-XXL~\citep{li2023blip}& 34.0 / 35.4 & 49.2 & 28.6 & 27.3 & 33.7 & 51.5 & 30.4  \\
\rowcolor{myopen} InstructBLIP-T5-XXL~\citep{Dai2023InstructBLIP} & 33.8 / 35.7 & 48.5 & 30.6 & 27.6 & 33.6 & 49.8 & 29.4  \\
\rowcolor{myopen} LLaVA-1.5~\citep{liu2023llava} & 33.6 / 36.4& 49.8 & 28.2 & 25.9 & 34.9 & 54.7 & 28.3  \\
\bottomrule
\end{tabular}
\caption{
Comparison with state-of-the-art methods on the MMMU dataset. *: model on leaderboard with unidentified reference. The models are ranked by overall test set scores (we report scores in a test/validation manner), including both \colorbox{myprop}{proprietary} and \colorbox{myopen}{open-sourced} models. We highlight our results with color green. \methodshort achieves SOTA performance in the open source category.
}
\label{tab:mmmu_comparison}
\end{table*}

\subsection{Gauging on Synthetic Data Quality and Hallucination}
Figure \ref{fig:winrate} presents cross-evaluation results with Gemini and GPT-4V, showing their increased preference for captions generated in later rounds of self-augmentation. We also provide evidence from an out-of-distribution benchmark and a human blend quality ranking, demonstrating that hallucinations do not increase in these later rounds.

\textbf{Left-out Benchmark Results.}
We first select the Visual Spatial Reasoning (VSR~\citep{liu2023visual}) benchmark which does not appear in our training set. This benchmark consists of triplets containing an image, a spatial-focused expression, (\eg, \textit{the cow is ahead of the person}), and a \textit{True} or \textit{False} label indicating its correctness. We observed reduced hallucinations on the VSR benchmark with more iterations of self-augmentation and have not yet reached a plateau, results are shown in Table~\ref{tab:vsr_results}.

\textbf{Human Judge Ranking Test.}
We further add a more rigorous human test that compares the (re-)captions of 200 randomly sampled images from 11 human evaluators (most are PhD students). These evaluators were tasked with determining which caption exhibits fewer hallucinations, without any knowledge of the sources of the captions. We calculated the win rates of the captions from later self-augmentation rounds against those from earlier rounds, as detailed in Table~\ref{tab:human_eval}. Human preference reflects a decrease in hallucinations with each additional round of self-augmentation, reaching saturation at Round-3. This observation aligns with the performance trends noted in Table~\ref{tab:iteration_rounds} and GPT-4V and Gemini judges in Figure~\ref{fig:winrate}. 

\begin{minipage}{\textwidth}
\begin{minipage}[h]{0.49\textwidth}
\centering
\scriptsize
\makeatletter\def\@captype{table}
\begin{tabular}{lcc}
    \toprule
    Model     & VSR \textit{random} (\%) $\uparrow$     & VSR \textit{zero-shot} (\%) $\uparrow$ \\
    \midrule
    VILA$_{0}$ & 73.5 & 63.1 \\
    VILA$_{1}$ & 75.1 & 64.8 \\
    VILA$_{2}$ & 76.8 & 65.8 \\
    VILA$_{3}$ & \textbf{77.4} & \textbf{66.4} \\
    \bottomrule
\end{tabular}
\caption{Accuracies on the VSR benchmar.}
\label{tab:vsr_results}
\end{minipage}
\begin{minipage}[h]{0.49\textwidth}
\centering
\scriptsize
\makeatletter\def\@captype{table}
\begin{tabular}{lc}
    \toprule
    Human Evaluation & Win Rate (\%) $\uparrow$\\
    \midrule
    Later Rounds vs Original Caption & 71.6 \\
    Later Rounds vs Round-1 & 54.6 \\
    Later Rounds vs Round-2& \textbf{55.6} \\
    Later Round \ \ vs Round-3 & 48.3 \\
    \bottomrule
\end{tabular}
\caption{Preferences for 200 images' captions.}
\label{tab:human_eval}
\end{minipage}
\end{minipage}

\subsection{Improved Data Quality Matters More than Increased Computation}
We next present additional ablations of \methodshort's self-augmentation loop in comparison to training additional epochs on the same data, with results shown in Table \ref{tab:more_epochs}. We can observe that simply scaling up epochs with coarse image/text pairs yields no performance boosts, despite the extra training costs. In contrast, enhancing the quality of data, as demonstrated in VILA, provides a more rewarding path.
\begin{table*}[h]
\setlength{\tabcolsep}{3pt}
\centering
\scriptsize
\begin{tabular}{lcccccc}
\toprule
Model Variation & GQA & SQA$^\text{I}$ & VQA$^\text{T}$ & POPE & MM-Vet & MMMU \\
\midrule
\rowcolor{black!10} VILA$_0$-\textit{Baseline} & 62.4 & 68.4 & 61.6 & 84.2 & 34.5 & 33.8 \\
Train one extra epoch & 62.5 & 68.7 & 61.9 & 84.0 & 34.4 & 33.9 \\
\rowcolor{black!10} VILA$_1$ & 63.2 & 71.0 & 62.5 & 84.6 & 34.8 & 35.8 \\
Train two extra epochs & 62.3 & 69.0 & 61.7 & 83.9 & 34.4 & 33.7 \\
\rowcolor{black!10} VILA$_2$ & 63.5 & 71.5 & 63.5 & 84.7 & 35.5 & 35.2 \\
\bottomrule
\end{tabular}
\caption{Comparison between training additional epochs \textit{on the same data} and training additional epochs \textit{with self-augmentation}. Models do not benefit from more computations on identical data.}
\label{tab:more_epochs}
\end{table*}
\subsection{Efficiency and Effectiveness of \methodshort}
\textbf{Cost Analysis -- Labeling.} Data quantity and quality are critical factors in model training. While \methodshort involves three rounds of recaptioning, it is still far more cost-efficient than traditional human re-labeling. For example, a standard re-labeling on Amazon Turk costs 36 USD per 1k images, while \methodshort costs only 0.12 USD per 1k images. The cost breakdown includes AWS pricing for H100 GPUs: USD 4.91 per hour for one H100, or USD 39.33 per hour for eight. With an inference speed of 10.6 images per second per H100, \methodshort processes around 38,340 images per hour, making it \textbf{300x} cheaper and significantly faster.

\textbf{Cost Analysis -- Training.} As shown in Table~\ref{tab:cost_analysis}, \methodshort achieves better accuracy with significantly less data (51M) compared to other works like MM1 (>2B) and Idefics2 (>600M). Even with three rounds of iteration, \methodshort remains more cost-effective in terms of training computation. Further, \methodshort is a \textit{one-time cost} that can be leveraged for training multiple models. Once recaptioning is complete, this data can be shared with the community, reducing the need for expensive data pipelines.
\begin{table*}[t]
\setlength{\tabcolsep}{2pt}
\centering
\resizebox{\textwidth}{!}{%
\begin{tabular}{lcccc|cccccccc}
\toprule
Method & LLM & VT & \# TPI & PT & $\text{VQA}^{\text{v2}}$ & SQA$^\text{I}$ & VQA$^\text{T}$ & MMB & SEED & LLaVA$^\text{W}$ & MM-Vet & MMMU \\
\midrule
MM1-7B-Chat~\cite{mckinzie2024mm1} & 7B & 300M & 720 & $>$2B & 82.8 & 72.6 & 72.8 & 72.3 & 64.0 & 81.5 & 42.1 & 35.6 \\
Idefics2-8B~\cite{laurençon2024mattersbuildingvisionlanguagemodels} & 8B & 400M & 320 & $>$600M & 81.2 & -- & 73.0 & \textbf{76.7} & -- & -- & -- & 37.7 \\
\rowcolor{myopen}VILA$^{2}$-8B (ours) & 8B & 400M & 196 & \textbf{51M} & \textbf{82.9} & \textbf{87.6} & \textbf{73.4} & 76.6 & \textbf{66.1} & \textbf{86.6} & \textbf{50.0} & \textbf{38.3} \\
\bottomrule
\end{tabular}
}
\caption{Comparison of multimodal methods across benchmarks, with different settings of large language model parameters (LLM), vision tower parameters (VT), number of tokens per image (\# TPI), and pre-training data size (PT).}
\label{tab:cost_analysis}
\end{table*}

\section{Related Work \& Limitations}

\textbf{Visual language models (VLM).} 
Visual language models have rapidly progressed in recent years ~\citep{Radford2021LearningTV,li2022blip,Dai2023InstructBLIP,liu2023llava,ye2024x,cheng2024spatialrgpt}.
The success mainly comes from pretraining visual and language models on the internet-scale data. 
Kosmos-2~\citep{peng2023kosmos} and PaLI-X~\citep{chen2023pali} largely scaled the pretraining data by pseudo-labeling bounding boxes from performant open-vocabulary object detectors (GLIP~\citep{li2022grounded} and OWL-v2~\citep{minderer2024scaling}, respectively). They examined that strong perception capabilities such as object detection and OCR translate to better high-level reasoning tasks like visual question-answering (VQA). 

\textbf{Contributions \& Novelty.} Our work expands the horizon of data-scaling through our self-augmenting paradigm. ShareGPT4V~\citep{chen2023sharegpt4v} applied a single round of recaptioning by distilling from GPT-4V. In contrast, we focus on a more general approach of \textit{using VLM to augment VLM itself} without relying on commercial APIs or distilling from larger models. We provide 1) a detailed analysis of self-augmentation, covering prompt templates, iteration rounds, saturation points; 2) a practical method that uses specialist-augmentation to continually improve; 3) curated datasets that can be reused for future research. Our solution efficiently enhances SOTA VLM performance without requiring extra data or expensive APIs for closed-source models. 

\textbf{Limitations}. Due to resource constraints, we concentrate on the design of a self-augmented data curation pipeline and verify the 7B, 8B, and 40B models with 51M data. Larger models (e.g., 70B and 405B) and more data (>0.5B pretrain data) can have the potential to lead to better VLM capabilities with self-augmenting abilities. We will address these aspects in future work.

\section{Conclusions}
This work has explored the techniques, insights, and benefits of using VLMs to self-improve its pretraining. We introduced two primary augmentation loops, one leveraging VLM's general captioning capacities and the other harnessing their strength in specialized visual tasks. We demonstrated the feasibility of three `free lunch' rounds for VLMs through self-bootstrapping, with further improvements via knowledge distillation from specialist VLMs. Our new \methodshort models demonstrate SOTA performances across a comprehensive set of benchmarks. Fruitful future directions include a deeper delve into the potential synergy between synthetic and real data to train stronger foundation models.

\bibliography{iclr2025_conference}
\bibliographystyle{iclr2025_conference}
\appendix

\clearpage\section{Appendix / supplemental material}

\subsection{Prompts for Specialist-augmentation}
\label{sec:appendix_A_1_1}
We use the following prompts during specialist- augmentation, 
\begin{itemize}
    \item Spatial Relations Understanding Specialist
    
    \mytt{"<image> Elaborate on the visual and narrative elements of the image in detail, with a focus on spatial relations."}
    \item Grounded Narration Specialist
    
    \mytt{"<image> Elaborate on the visual and narrative elements in the image, and specify their location with [xmin,ymin,xmax,ymax]."}
    \item OCR Specialist
    
    \mytt{"<image> Your task is to recognize and describe the text in the image. Please provide a brief description that focuses on the textual content."}
\end{itemize}

\subsection{Specialist Acquisition}
\label{sec:appendix_A_2}

\begin{figure}[ht]
    \centering
    \includegraphics[width=1.0\linewidth]{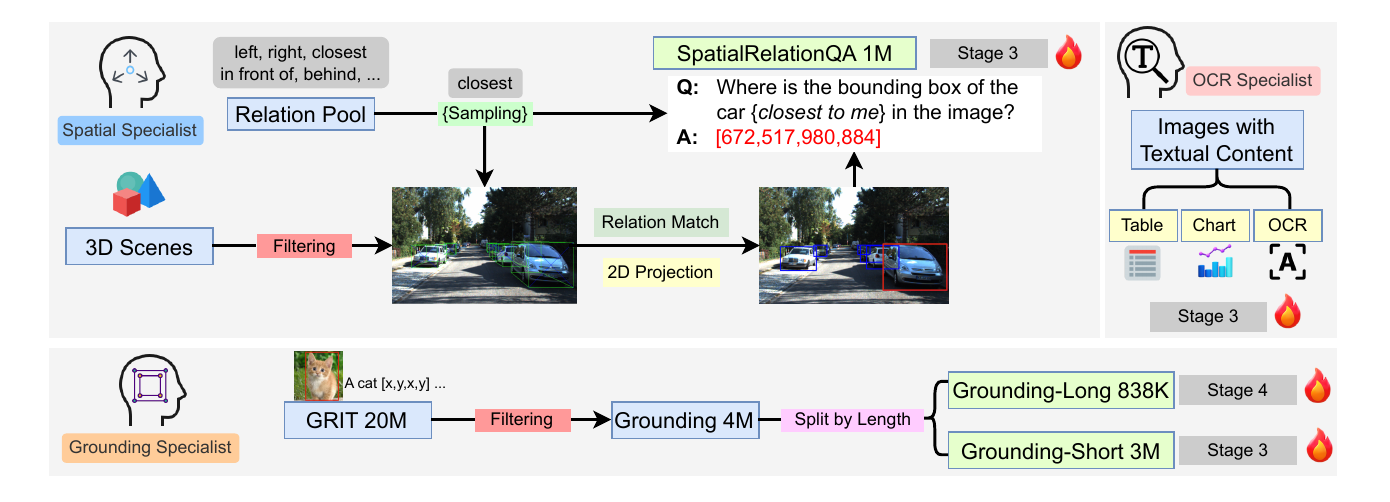}
    \caption{
         \methodshort Specialist VLM Acquisition Pipeline. We gather task-specific knowledge from public datasets, followed by filtering noisy samples using rule-based strategies. We then train the specialist VLMs from pretrained checkpoints, employing different data blends and training strategies.
    }
    \label{fig:speciality_acquisition}
\end{figure}



Specialty and expertise can be obtained via gathering existing data from the open-source community, human labeling, and annotating by domain-specific models, e.g. OCR models for text recognition and detection models for bounding box prediction. We also experimented with using open-world detectors like OWLv2\citep{minderer2024scaling} to automatically label bounding boxes, VLMs to generate detailed captions, and LLMs such as Llama3-70B-Instruct to merge the information for the grounded narration specialist. However, we found that language models introduced more hallucinations into the merged grounded narration. This is because many different detection labels can share the same meaning and refer to the same instance, making it difficult for the language model to perform the bipartite matching between bounding boxes and their text correspondence.

\subsection{SFT Data}
\label{appendix:sft_data}
We use two different datasets for our experiments: a 1.8M sample dataset for exploratory experiments and a 5.9M sample dataset for state-of-the-art experiments.

\begin{itemize}
    \item \textbf{1.8M SFT Blend: }This dataset includes samples from the following sources: LLaVA-SFT, MSR-VTT, TextCaps, Image Paragraph Captioning, CLEVR, NLVR, VisualMRC, ActivityNet-QA, iVQA, MSRVTT-QA, MSVD-QA, DVQA, OCRVQA, ST-VQA, ViQuAE, VQAv2-train, Visual Dialog, GQA-train, FLAN-1M.
    \item \textbf{5.9M SFT Blend: }This dataset comprises all the datasets listed in the following table:
\end{itemize}

\begin{table}[h]
\centering
\scalebox{0.9}{
\begin{tabular}{c|c}
\toprule
Categories & Datasets\\ \midrule
Hybrid     & LLaVA-SFT, The Cauldron (subset)\\\midrule
Captioning & \makecell{MSR-VTT~\citep{xu2016msr}, TextCaps, LLaVAR, \\Image Paragraph Captioning~\citep{krause2017hierarchical}, ShareGPT4V-100K} \\ \midrule
Reasoning  & CLEVR~\citep{johnson2017clevr}, NLVR, VisualMRC~\citep{tanaka2021visualmrc} \\\midrule
Multi-images & \makecell{ActivityNet-QA~\citep{yu2019activitynet}, VQAv2-train, \\iVQA~\citep{yang2021just}, MSRVTT-QA, STEM-QA~\citep{shen2024measuringvisionlanguagestemskills}} \\ \midrule
OCR & \makecell{DVQA, OCRVQA, ST-VQA~\citep{biten2019scene}, \\SynthDoG-en,  TextOCR-GPT4V, ArxivQA}\\ \midrule
World Knowledge & \makecell{WIT~\citep{Srinivasan_2021}} \\ \midrule
General VQA        & \makecell{ScienceQA-train, VQAv2-train, \\ViQuAE~\citep{lerner2022viquae}, Visual Dialog~\citep{das2017visualdialog}, \\GQA-train~\citep{hudson2019gqa}, SHERLOCK~\citep{hessel2022abductionsherlockholmesdataset}, \\Geo170K~\citep{gao2023gllavasolvinggeometricproblem}, MMC-Instruction~\citep{liu2024mmcadvancingmultimodalchart}, \\LRV-Instruction~\citep{liu2024mitigatinghallucinationlargemultimodal}, RefCOCO-train~\citep{yu2016modelingcontextreferringexpressions}}\\ \midrule
Text-only & \makecell{FLAN-1M~\citep{wei2022finetunedlanguagemodelszeroshot}, MathInstruct~\citep{yue2023mammothbuildingmathgeneralist}, \\Dolly~\citep{DatabricksBlog2023DollyV2}, GSM8K-ScRel-SFT~\citep{yuan2023scalingrelationshiplearningmathematical}} \\
\bottomrule
\end{tabular}
}
\vspace{0.1in}
\caption{Data mixture for the SFT stage.}
\label{tab:appendix-data}
\end{table}

\subsection{Specialist Data}
\label{appendix:specialist_data}
We integrated specialty data with high-quality image captioning datasets and diverse instruction finetuning datasets, ensuring the models retain their narrative and instruction-following abilities while acquiring task-specific knowledge.

\begin{enumerate}
    \item \textbf{Spatial Specialists.} We continued training the specialist from the stage 2, ALLaVA caption~\citep{chen2024allavaharnessinggpt4vsynthesizeddata}, and GPT-4V caption from ShareGPT4V~\citep{chen2023sharegpt4v}.
    \item \textbf{Grounding Specialist.} We split the cleaned 4M grounded narration into \textit{Grounding-Short} 3M and \textit{Grounding-Long} 838K for a two-stage training process. In stage 3, we combined \textit{Grounding-Short} 3M with ALLaVA caption~\citep{chen2024allavaharnessinggpt4vsynthesizeddata} to adapt to new tasks of grounded narration while maintaining the narrative ability. In stage 4, we combine \textit{Grounding-Long} 838K with Shikra GPT-4~\citep{chen2023shikra}, Visual7W~\citep{zhu2016visual7w}, LLaVA-SFT, and 100K GPT-4V captions from ShareGPT4V to sustain both narrative and instruction following capacities.
    \item \textbf{OCR Specialist.} We trained our OCR specialist with various internet datasets focused on text recognition, understanding, and reasoning, including LLaVA-SFT, TextOCR-GPT4V~\citep{textocr-gpt4v}, SynthDoG-En~\citep{kim2022ocr}, OCRVQA~\citep{Mishra2019}, TextCaps~\citep{sidorov2020textcaps}, ArxivQA~\citep{li2024multimodal}, DocVQA~\citep{kafle2018dvqa}, AI2D~\citep{kembhavi2016diagram}, ChartQA~\citep{masry2022chartqa}, LLaVAR~\citep{zhang2023llavar} and 35 OCR-related datasets from The Cauldron~\citep{laurençon2024mattersbuildingvisionlanguagemodels}.
\end{enumerate}

\subsection{Training Detail}
\label{appendix:training_detail}
We adjust our training strategies akin to varying language model sizes for training cost considerations. We next elaborate on the details.

\subsubsection{7B \& 8B \& 13B Models}
We divide the entire training process of 7B\&8B\&13B models into three sub-stages.
\begin{itemize}
    \item \textbf{Stage 1: Alignment Stage}.We train only the multi-modal projector using 595K image-text pairs, as mentioned in LLaVA, to achieve the initial alignment between the two modalities.
    \item \textbf{Stage 2: Pretraining Stage}. We gather a total of 51 million images, consisting of 25 million image-text pairs with the highest CLIP scores from COYO-700M, 25 million images in an interleaved image-text format from the MMC4-Core subset, and 1 million images with detailed captions from ShareGPT4V-Pretrain. During this stage, we unfreeze both the multi-modal projector and the language model to enhance comprehension of the diverse multi-modal training data. \textbf{We use the augmented data here to replace the original COYO captions.}
    \item \textbf{Stage 3: Supervised Finetuning Stage}. After stage 2, we collect diverse visual question-answer pairs and unfreeze all parameters to finetune the model for general-purpose VQA capacities. 
\end{itemize}

\subsubsection{40B Model}
For the \methodshort-40B model, we skip the cost-intensive stage 2 and train the model with 7.5 million images randomly sampled from the 25 million COYO subset pairing with various caption sources: 2.5 million with original COYO captions, 2.5 million with \vila{3} re-captioned descriptions, and 2.5 million with \vila{3} spatial specialist re-captioned descriptions. Both the multi-modal projector and the language model remain unfrozen. Note that adding interleaved data, such as MMC4, can further boost the performance and we leave this potential to future work. A detailed profiling of 40B performances over benchmarks is also included as Table~\ref{tab:sota_comparison_appendix}.

\begin{table*}[t]
\setlength{\tabcolsep}{3pt}
\centering
\scalebox{0.65}{
\begin{tabular}{l ll | lllll | lllll }
\toprule
Method & LLM & Res. & VQA$^\text{v2}$ & GQA & VizWiz &  SQA$^\text{I}$ & VQA$^\text{T}$& MMB & MMB$^\text{CN}$ & SEED & LLaVA$^\text{W}$ & MM-Vet \\
\midrule
\methodshort-8B (ours) & Llama 3-8B & 384& 82.9 & 64.1 & 64.3 & 87.6 & 73.4 & 76.6 & 71.7 & 66.1 & 86.6 & 50.0 \\
\methodshort-40B (ours) & Yi-34B & 448& 85.1 & 64.7 & 62.2  & 93.2 & 75.9 & 83.9 & 82.9 & 77.0 & 93.6 & 53.4 \\
\bottomrule
\end{tabular}
}
\caption{
Improvements from 8B to 40B checkpoints on 10 visual-language benchmarks.
}
\label{tab:sota_comparison_appendix}
\end{table*}

\subsection{Hyperparameters}
We use a universal batch size of 1024, a cosine decay learning rate schedule, a 0.03 learning rate warmup ratio, no weight decay, and AdamW as the optimizer for stable training, and details are expanded in Table~\ref{tab:appendix-recipe}. All trainings are conducted with 128 A100 GPUs.

\begin{table}[h]
\centering
\scalebox{0.9}{
\begin{tabular}{l|cccc}
\toprule
\textbf{Hyperparameter}  & \textbf{Stage 1} & \textbf{Stage 2} & \textbf{Stage 3} & \textbf{Stage 4} \\
\midrule
batch size      &    1024 & 1024 & 1024 & 1024     \\
learning rate (lr) &    1e-3     & 5e-5     & 2e-5     & 1e-4     \\
lr schedule     & cosine & cosine & cosine & cosine     \\
lr warmup ratio & 0.03 & 0.03 & 0.03 & 0.03             \\
weight decay    & 0 & 0 & 0 & 0                 \\
epoch           & 1 & 1 & 1 & 1                 \\
optimizer       & AdamW  & AdamW & AdamW & AdamW         \\
DeepSpeed stage & stage2  & stage3 & stage3 & stage3  \\
\bottomrule
\end{tabular}
}
\vspace{0.1in}
\caption{The detailed training setup for \methodshort and the hyper-parameters across the training stages. }
\label{tab:appendix-recipe}
\end{table}



\subsection{Additional Re-caption Comparisons}

\begin{figure}[H]
    \centering
    \includegraphics[width=1.0\linewidth]{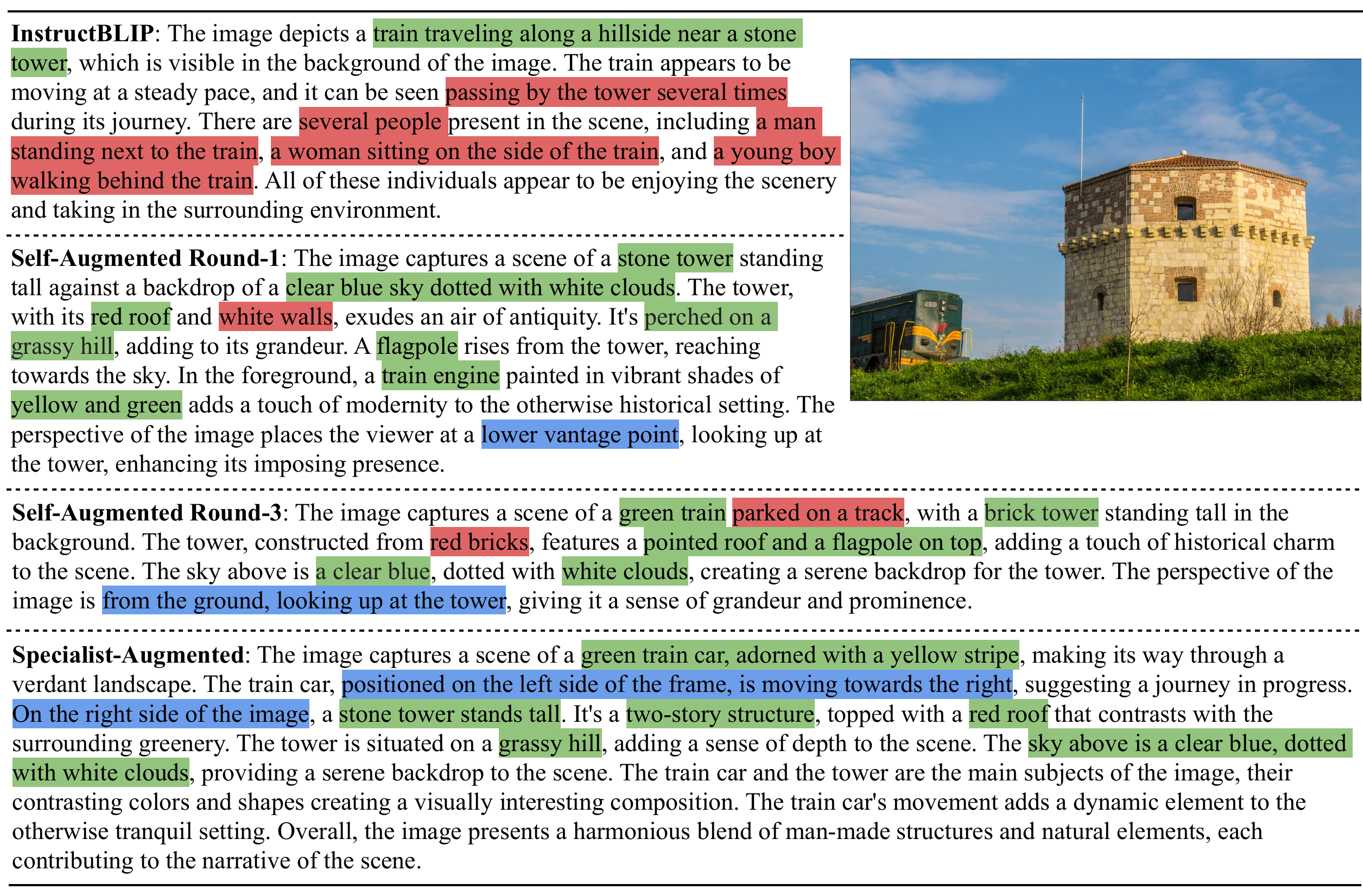}
    \caption{
        More examples of comparison among captions from generation rounds of \methodshort and the widely used caption baseline of InstructBLIP~\citep{Dai2023InstructBLIP}.
    }
    \label{fig:more_examples1}
\end{figure}
\begin{figure}[H]
    \centering
    \includegraphics[width=1.0\linewidth]{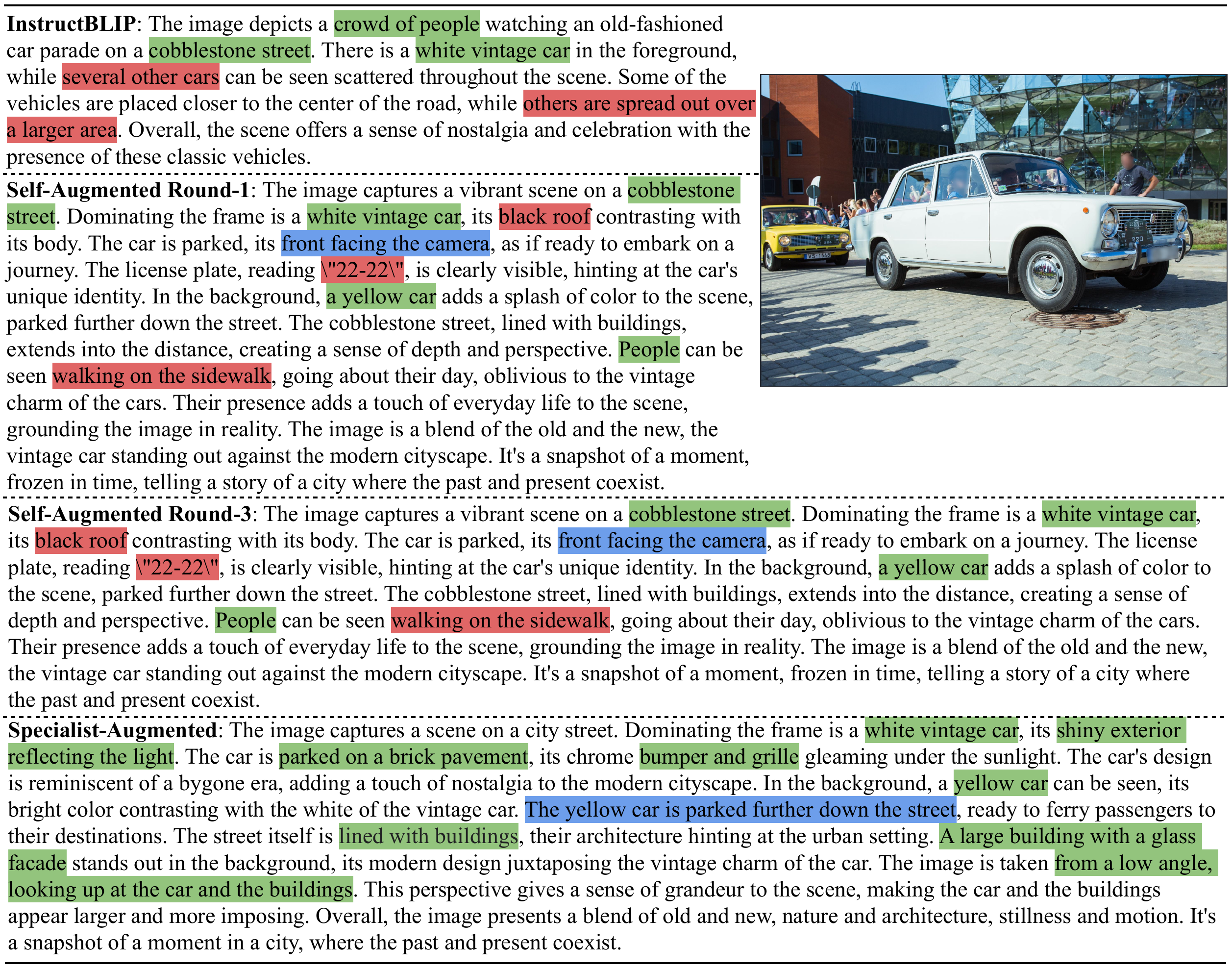}
    \caption{
        More examples of comparison among captions from generation rounds of \methodshort and the widely used caption baseline of InstructBLIP~\citep{Dai2023InstructBLIP}.
    }
    \label{fig:more_examples2}
\end{figure}
\begin{figure}[H]
    \centering
    \includegraphics[width=1.0\linewidth]{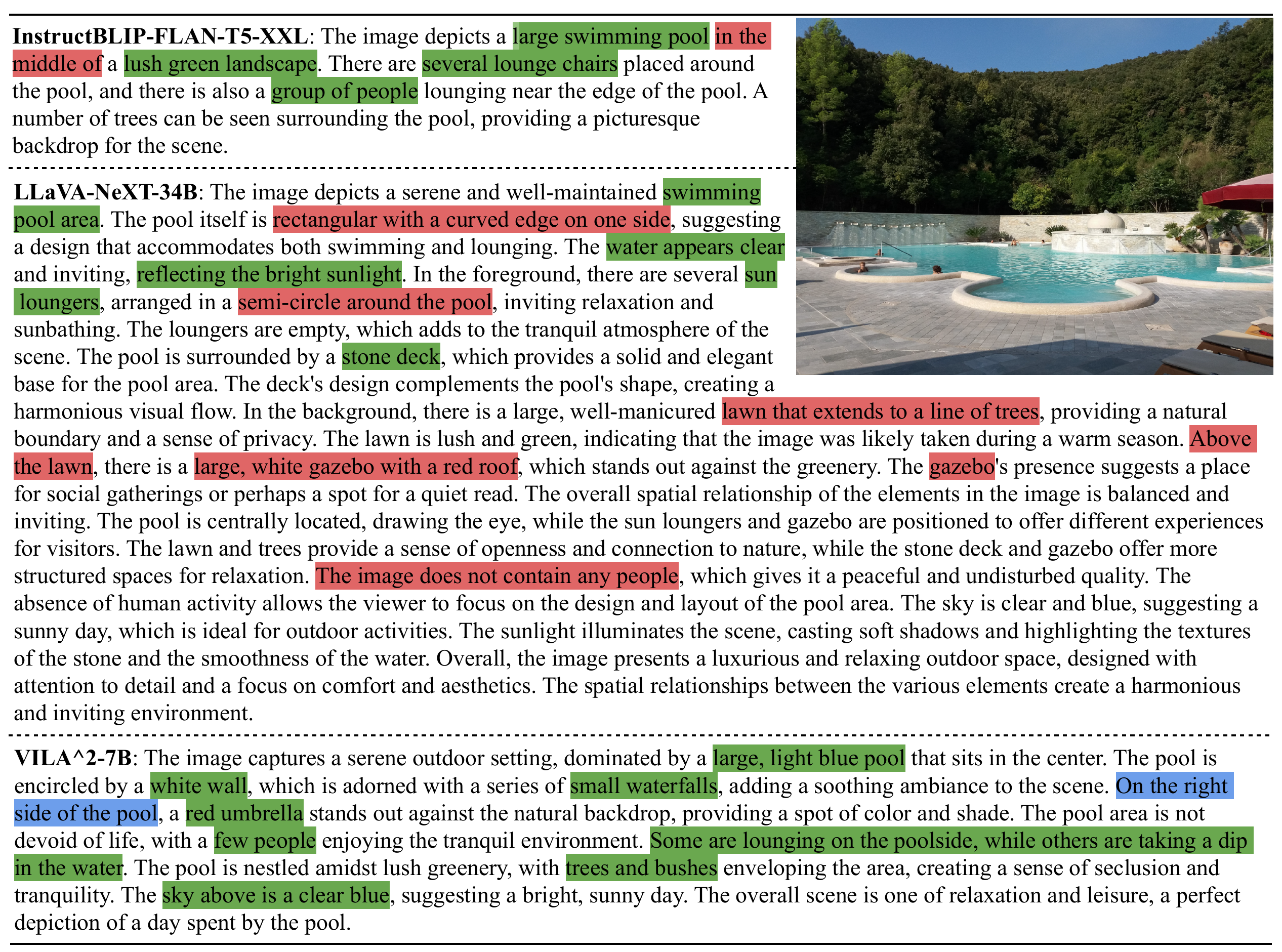}
    \caption{
    More examples of comparison among captions from generation rounds of \methodshort and an additional caption baseline of LLaVA-NeXT~\citep{liu2024llavanext}. LLaVA-NeXT-34B tends to generate longer caption with more hallucinations.
    }
    \label{fig:more_examples5}
\end{figure}

We provide additional \methodshort recaptioned examples from the SAM dataset comparing the baseline captions from alternative methods, \eg, the widely adopted InstructBLIP Flan-T5 XXL~\citep{Dai2023InstructBLIP} and LLaVA-NeXT-34B~\citep{liu2024llavanext}, with captions generated through various rounds of \methodshort's intermediate models in Figure~\ref{fig:more_examples1} through Figure~\ref{fig:more_examples5}. The correct facts are marked with \textcolor{nvgreen}{green}, hallucinations are marked with \textcolor{red}{red} and spatial related information are marked with \textcolor{blue}{blue}. From these examples, we can see that while the baseline caption is descriptive, it contains more hallucinations. As the re-captioning and training cycle progresses, the model tends to generate captions with more details and fewer hallucinations. Additionally, the spatial specialist introduces more spatial information into the recaptions, as expected.

\end{document}